\documentclass[runningheads]{llncs}
\pdfoutput=1 

\usepackage[utf8]{inputenc}
\usepackage{hyperref}
\usepackage[misc]{ifsym}
\usepackage{graphicx}
\usepackage{wrapfig}
\usepackage{verbatim}
\usepackage{paralist, amsmath, color, enumerate, framed}
\usepackage{subcaption}
\usepackage{todonotes}
\usepackage{soul}
\usepackage{multirow}
\usepackage{tabularx}
\usepackage{array}

\usepackage{listings}
\usepackage{changepage}
\usepackage{makecell}
\usepackage{booktabs}
\newcolumntype{P}[1]{>{\centering\arraybackslash}m{#1}}
\title{Knowledge Conceptualization Impacts RAG Efficacy}
\author{%
Chris Davis Jaldi\orcidID{0009-0000-2287-1198} \and
Anmol Saini\orcidID{0009-0000-1735-2377} \and
Elham Ghiasi\orcidID{0009-0009-1201-2029} \and
O. Divine Eziolise \and
Cogan Shimizu\orcidID{0000-0003-4283-8701}}
\institute{Wright State University, USA \\
\email{\{jaldi.2,saini.25,ghiasi.2,eziolise.5,cogan.shimizu\}@wright.edu}
}
\authorrunning{Jaldi, C. D., et al.}
\begin{document}
% Do frontmatter
\maketitle
% Do abstract
\begin{abstract}
    Explainability and interpretability are cornerstones of frontier and next-generation artificial intelligence (AI) systems. This is especially true in recent systems, such as large language models (LLMs), and more broadly, generative AI. On the other hand, adaptability to new domains, contexts, or scenarios is also an important aspect for a successful system. As such, we are particularly interested in how we can merge these two efforts, that is, investigating the design of transferable and interpretable neurosymbolic AI systems. Specifically, we focus on a class of systems referred to as ``Agentic Retrieval-Augmented Generation'' systems, which actively select, interpret, and query knowledge sources in response to natural language prompts. In this paper, we systematically evaluate how different conceptualizations and representations of knowledge, particularly the structure and complexity, impact an AI agent (in this case, an LLM) in effectively querying a triplestore. We report our results, which show that there are impacts from both approaches, and we discuss their impact and implications. 
\end{abstract}
%%%%%%%%%%%%%%%%%%%%%%%%%%%%%%%%%%%%%%%%%%%%%%%%%%%%%%%%%%%%%%%%%
%%%%%%%%%%%%%%%%%%%%%%%%%%%%%%%%%%%%%%%%%%%%%%%%%%%%%%%%%%%%%%%%%
\section{Introduction}
\label{sec:intro}
%%%%%
% Hook and introduce some basic hook concepts of the paper
Recent advances in large language models (LLMs) and the exploration of their abilities across diverse tasks, ranging from simple language processing and understanding to near-reasoning and recommender systems, have clearly positioned them well and not as simple generative AI models \cite{huang2022towards}. While these large models are capable of broad versatile actions, their long-standing limitations, accompanied by the heavy operational costs associated with them, persist \cite{banerjee2024llms}. These constraints, alongside unstopped advancements, have catalyzed the development of Retrieval-Augmented Generation (RAG) systems \cite{lewis2020retrieval}, which dynamically integrate and leverage both structured and unstructured external knowledge bases into the generation process. These RAG systems, especially when augmented with symbolic structured knowledge, retrieve relevant information with response to user prompt queries, offering a way for a hybrid Neurosymbolic AI (NeSy) \cite{hitzler2022neuro} architecture that combines the natural language capabilities of LLMs with factual precision of longitudinally curated knowledge sources.

% Positions KGs, ontology, KE, etc, and their importance.
Among various structured knowledge assets available, knowledge graphs (KGs), with graph-based semantics and expressive logical formalizations, emerged as a critical paradigm for supporting accurate, interpretable, and semantically rich knowledge modeling and retrieval \cite{kgs-hitzler,kgs-hogan,kgs-noy}. However, though there are several effective methods, including black-box or hard retrieval\footnote{A method which exhaustively retrieves all triples, places them into a vector-like space, and retrieves based on k-nearest neighbor search}, the integration of KGs into neural systems with RAG approaches remains non-trivial. The design of a KG's schema -- consisting of the conceptualization of entities, relationships, and their constraints -- can fundamentally limit a system's capacity for querying and navigating knowledge effectively. Thus, this issue falls squarely within the intricacies of knowledge conceptualization, and it is a core concern for ontology engineering and semantic web research, where the focus and concerns usually lie in leveraging and structuring domain-specific knowledge for consistent interpretations and computational reasoning tasks \cite{gruber}.

%%%%%%%%%%
Despite the increased interest in hybrid NeSy systems, the effects and influence of schema and ontology representations across various complexities on a RAG system's performance is a research concern area that has been explored by few, if any. Many existing RAG pipelines usually treat KGs as a passive data source, with little to no attention paid to their internal conceptual organization and effective leveraging \cite{soman2024biomedical,zhang2025survey}. The advent of Agentic RAG systems, a collection of neural systems that communicate in chains for dynamic retrieval, opens up an avenue for optimal intelligent querying of KGs. Bridging these advances paves the way to explore the ability of LLMs to query over KGs using various querying languages like SPARQL \cite{meyer2024assessing,site:nl-to-sparql-llm,sequeda2024benchmark}. Yet, research on assessing and leveraging the capability of AI to understand and facilitate effective querying remains an underexplored area, regardless of the structural representation methods, which compels for a shift in paradigm and raises important questions about whether LLMs can meaningfully internalize, interpret, and leverage these conceptual structures.

%%%%%
\begin{figure}[t]
    \centering
    \includegraphics[width=1\textwidth]{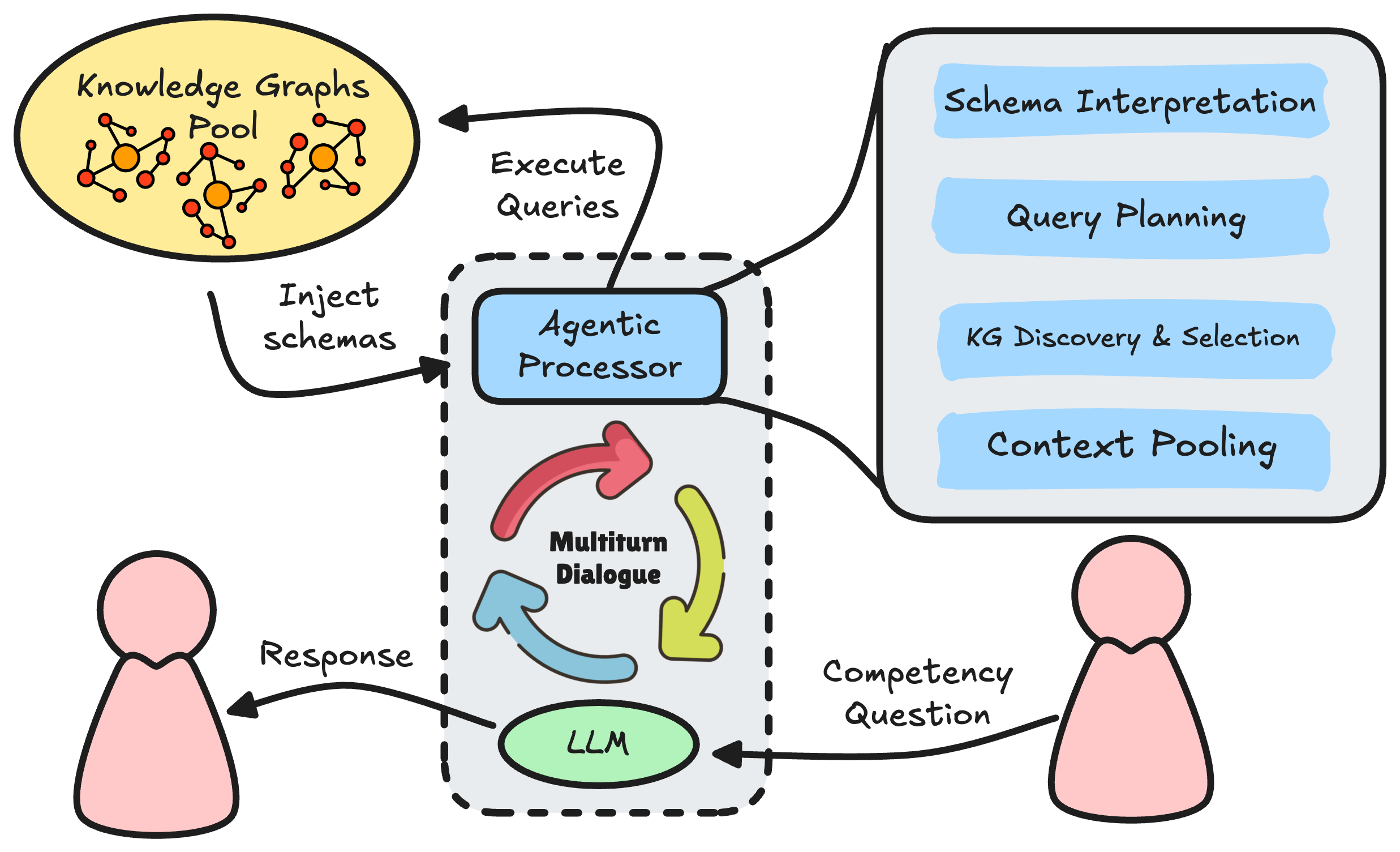}
    \caption{This figure overall demonstrates the positioning of our work in agentic graph-RAG systems. Specifically, we are focused on how the manner of schema injection into the prompts impacts the LLM's ability to construct semantically and syntactically correct SPARQL queries.}
    \label{fig:positioning}
\end{figure}
%%%%%

We position an \emph{agentic} graph-RAG system (see Figure~\ref{fig:positioning}) in which specific AI agents autonomously select knowledge sources (in our case, KGs), reason over partial schema extracts, and generate SPARQL queries for retrieval of relevant content. The current research advances our understanding of important aspects of the LLM-agents' ability to understand a knowledge base's schema, as well as leveraging the ontological rules defined to comprehensively retrieve all insights about our data and its relationships, thus addressing the very critical problem of how we can move forward.

As such, we investigate the following research question:
\begin{enumerate}[\bf RQ1.]
  \item How does the schema (ontology) complexity and its representation in KGs affect the efficacy of \emph{agentic} RAG systems?
\end{enumerate}

To address this, we systematically explore and evaluate this process across KGs, varying how the schema is represented -- from simple triple representation to description logic axioms \cite{baader2008description} serialized in Manchester Syntax \cite{manchester-tr} -- and the complexity of its conceptualization (as a function of reification). The study also spans two domains, to account for domain-specific bias. We generate queries based on real-world competency questions (CQs) \cite{bezerra2013evaluating} taken from the original projects. Specifically, our work addresses the following two hypotheses.
\begin{enumerate}[\bf H1.]
  \item \textbf{Schema Complexity Matters:} The complexity of the ontology, measured primarily through reification, affects the model's ability to produce semantically and syntactically correct queries generated from CQs. We expect an inverse correlation between complexity and performance.
  
  \item \textbf{Schema Representation Matters:} The manner in which the ontology is represented to the LLM affects its ability to generate semantically and syntactically correct queries generated from CQs. We expect that when a simpler representation is used, the model will generally perform better than otherwise.
\end{enumerate}

By critically comparing the queries generated and their projected retrieval quality across our different schema representations, we explore how different knowledge conceptualizations, captured through schema design, mediate the RAG systems' effectiveness. This is a timely, important, and emergent research area, where it is critical to understand how to effectively leverage ontological features for effectively querying knowledge bases through active retrieval, as opposed to treating them as passive data sources. This exploration has potential implications for knowledge structure design and representation in the emerging landscape of NeSy systems. These findings also contribute to broader discussions on semantic interoperability with AI systems, offering practical guidelines for ontology modeling, which further support LLM-driven systems \cite{norouzi2025ontology,shimizu2025accelerating}.

% Directory Information
The rest of this paper is organized as follows.
In Section~\ref{sec:rel}, we briefly discuss some foundational concepts around the current state-of-the-art, as well as similar efforts. 
Section~\ref{sec:meth} provides the methodological details on the experiment design. 
We present our results in Section~\ref{sec:results}, followed by a discussion of their implications in Section~\ref{sec:disc}.
Finally, we conclude with Section~\ref{sec:conc}, alongside future directions.

%%%%%%%%%%%%%%%%%%%%%%%%%%%%%%%%%%%%%%%%%%%%%%%%%%%%%%%%%%%%%%%%%
\section{Related Work}
\label{sec:rel}
%%%%%

Integrating an external knowledge base as a RAG target, and especially a KG~\cite{rajabi2024knowledge}, has had growing attention over recent times~\cite{shimizu2025accelerating}, and thus home to recent innovation. As such, we have identified several efforts that have acted as a foundation for, or otherwise influenced, our approach. While RAG systems demonstrated value to improve the factual accuracy and domain alignment of LLM outputs, the role of knowledge design and representation, schema complexity, and semantic modeling choices in shaping retrieval and generation quality is still underexplored. In the following, we briefly review prior work and critically compare our own work to a few such efforts.  

Recent surveys categorized RAG systems by their ability to utilize external data to improve LLM grounding and reduce hallucinations~\cite{zhao2024retrieval}. These works differentiate RAG tasks between simple fact retrieval and complex rationale synthesis, sometimes involving an agentic system that performs hard task decomposition using approaches similar to the LADDER framework~\cite{simonds2025ladder}, and note that retrieval strategies must align with the complexity of the said task. While much of this focus was on unstructured text retrieval, several studies explored structured knowledge integration, especially for Question Answering or QA-RAG, as some call it~\cite{mansurova2024qa}. Some demonstrate the value of leveraging minimal schema with embedding-based context pruning to improve token efficiency while maintaining semantic grounding~\cite{soman2024biomedical}. However, these methods highlight the importance of schema-driven context reduction but largely treat schema as static background metadata rather than as an active design variable that influences system behavior. Additionally, we know that structure impacts graph embeddings, or even KG embeddings in the way they are translated across the space~\cite{christouexperiments,dave2023towards,dave2024towards}, but assessing if the same would have an effect on user querying is an unanswered question.

Several prior works have explored querying as a bridge between natural language and structured KGs. While LLMs showed capabilities to generate valid SPARQL queries~\cite{meyer2024assessing,sequeda2024benchmark}, ensuring that semantic correctness and schema conformity remain is a challenge. Some others proposed staged pipelines that precondition LLMs on schema fragment selection to improve query performance on large-scale KGs~\cite{yang2023llm}, while others extended these pipelines to federated querying across multiple KGs, using representations like VoID\footnote{Vocabulary of Interlinked Datasets \url{w3.org/TR/void}} and Shape Expressions (ShEx)~\cite{emonet2024llm}. These approaches highlight the potential value of schema validation but do not fully address how schema complexity itself impacts performance. One-shot prompting strategies that combine several schema fragments, a sample SPARQL query, and required questions have also been proposed. While they were demonstrating promising results on familiar datasets, these methods often struggle to generalize when schema structures become more complex or unfamiliar~\cite{kovriguina2023sparqlgen}.

Despite these advancements, the impact of schema representation and structural complexity in RAG performance remains under-characterized. Existing work generally focuses on improving prompting techniques or reducing token size, without systematically comparing different schema modeling strategies. Moreover, the scaling challenges introduced by LLM context window limits are rarely explored in relation to schema size and expressivity. Practical use in the real world requires balancing schema completeness with prompting feasibility, but few to no studies offer valuable insights into navigating this trade-off effectively.
Additionally, most studies do not account for:
\begin{inparaenum}[\bf(a)]
    \item The interaction between schema complexity and LLM reasoning demands,
    \item The impact of human-readable versus machine-oriented schema representations,
    \item The scaling limits imposed by context window constraints, and
    \item The potential for prompt engineering to mitigate schema-induced hallucinations.
\end{inparaenum}

Our work gives insights into these gaps by systematically evaluating schema representation methods and complexity levels across multiple KGs and competency question types. By analyzing both axiomatized and simplified schema representations, and with implications for generalization and scalability, we contribute to how knowledge conceptualization directly impacts RAG system efficacy.

%%%%%%%%%%%%%%%%%%%%%%%%%%%%%%%%%%%%%%%%%%%%%%%%%%%%%%%%%%%%%%%%%
\section{Methodology}
\label{sec:meth}
\begin{figure}[t]
    \centering
    \includegraphics[width=1\textwidth]{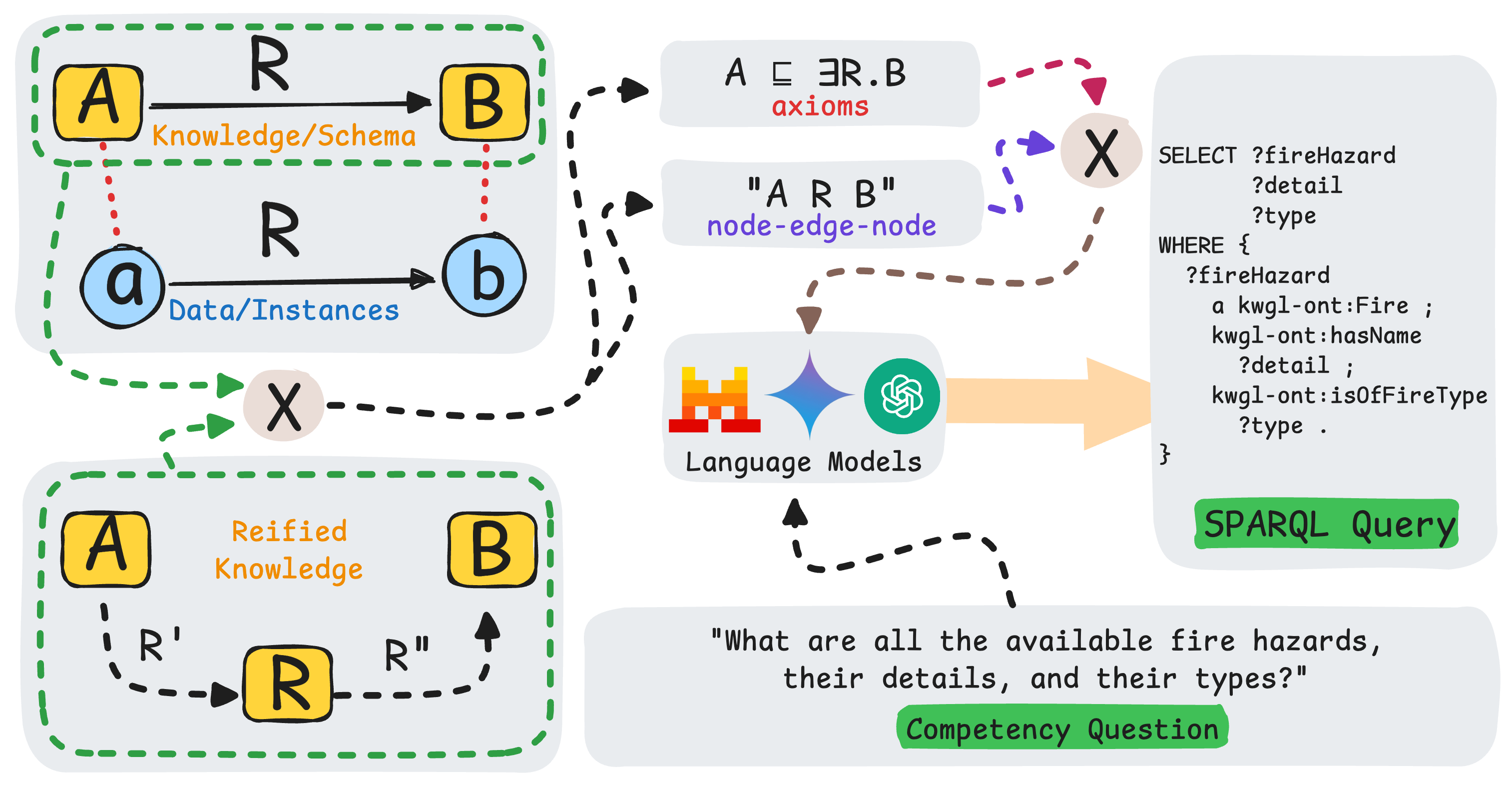}
    \caption{This figure shows a graphical depiction of our methodology. We have taken four schemas: KnowWhereGraph, KWG-Lite, the Enslaved.org Ontology, and the Enslaved.org Wikibase. These are expressed axiomatically or conceptually to an LLM alongside a CQ. The correctness of the generated SPARQL query is finally assessed.} % redo
    \label{fig:exps}
\end{figure}

%%%%%
Our approach to conducting the experiment is detailed in this section. Specifically, it entailed first identifying various knowledge sources and their representations, followed by converting said representations into appropriate formats. Providing these representations, along with CQs of varying complexity, to LLMs as context, we sought to gauge their ability to generate SPARQL queries that accurately address the CQs. We followed a 2 (knowledge sources) × 2 (representations) × 2 (formats) experimental design. Finally, we evaluated the generated queries (see Figure~\ref{fig:exps}). All of our work and results are publicly available~\cite{site:git-repo}.

%%%%%
%%%%%

%%%%%%%%%%%%%%%%%%%%%%%%%%%%%%%%%%%%%%%%%%%%%%%%%%%%%%%%%%%%%%%%%
\subsection{Knowledge Sources}
\label{subsec:ds-kgs}
%%%%%
In selecting the data sources for this experiment, we prioritized KGs with multiple, already developed representations across diverse domains, so as to provide a means of comparison for the SPARQL query results and analyze how the data semantics would be modeled in the generated queries. Opting for publicly available and open resources, we selected the KGs (ontologies) from the KnowWhereGraph and Enslaved.Org Hub projects, which had the added benefit of in-house experience with this data.

The KnowWhereGraph (KWG) is one of the largest, publicly available geospatial KGs \cite{site:kwg,zhu2025knowwheregraph}. It integrates numerous datasets on factors like disasters, environmental observations, demographics, soils, and health to support applications in the realms of humanitarian relief, food, and agriculture. Thus, in conjunction with a series of developed tools that aid in the retrieval of useful data for users of various specialties and levels of technical skill, KWG offers a query endpoint website for the purpose of assisting domain scientists with decision-making.

The Enslaved.Org Hub \cite{site:enslaved} KG, containing information about the people of the historical slave trade, is a resource that unifies disparate collections of human-centric data, while also establishing a standard of best practices for data curation in the domain \cite{shimizu2020enslaved}. Thus, the efforts behind this project resulted in the development of a website and formally described ontology for querying pertinent data with the goal of preserving the slave trade's legacy and informing dialogue on the related moral and socioeconomic factors in the present.

%%%%%%%%%%%%%%%%%%%%%%%%%%%%%%%%%%%%%%%%%%%%%%%%%%%%%%%%%%%%%%%%%
\subsection{Schema Representation Strategies}
\label{subsec:schema-rep}
%%%%%
With the determination of KGs with multiple representations being a cornerstone of our experimental strategy, we observed two representations each of the KWG and Enslaved schemas. For each of these four representations, we established two avenues of providing the schema to an LLM: a node-edge-node (NEN) triples representation and an axiomatic representation, delivered as text, comma-separated value (CSV), or markdown files. The NEN representation expresses the schema in the form of triples, describing connections between entities via relationships and mirroring a graph structure in textual format. The axiomatic representation, however, captures the schema using formal logical statements while defining classes, properties, and other constraints to specify the ontology's semantics in a more rule-based, declarative manner.

Based on the availability and type of schema diagram files, we either programmatically generated the NEN representations from GraphML files (i.e., KWG) or manually curated them from the figures (i.e., Enslaved). We chose Manchester Syntax \cite{horridge2006manchester} for axiomatically capturing the ontology due to its natural language-like structure and the presumed proficiency of an LLM in parsing it over alternative representations~\cite{tam2024let}. However, we observed limitations with this design choice, as it does not fully capture the pertinent semantics. Additionally, we note the information loss when saving a Web Ontology Language (OWL) or Terse RDF Triple Language (TTL) file in Manchester Syntax using Protégé~\cite{site:protege}, an issue identified during the later phases of experimentation. For the alternative representations, we also found that the axioms were not readily available, prompting us to programmatically generate the scoped domain and range based on the corresponding NEN representations. Regardless, we conducted eight individual experiments and analyzed their results both independently and comparatively. We delve into specific details of the KWG and Enslaved representations here.

%%%%%%%%%%%%%%%%%%%%%%%%%%%%%%%%%%%%%%%%%%%%%%%%%%%%%%%%%%%%%%%%%
\noindent\textbf{KWG Representations}
% \label{subsubsec:kwg-rep}
%%%%%
The KWG KG \cite{zhu2025knowwheregraph} uses a modular approach in its modeling. It incorporates modules on characteristics like physical and political geographical features, hazards, socioeconomic factors, and their relations to each other. The alternative KWG representation, KWG-Lite, acts as a simplified version of KWG, predominantly modeling the same concepts and data but with fewer and less complex relationships \cite{shimizu2023knowwheregraph}. As a result, it also generally requires fewer hops to reach desired data from an arbitrary starting point.

%%%%%%%%%%%%%%%%%%%%%%%%%%%%%%%%%%%%%%%%%%%%%%%%%%%%%%%%%%%%%%%%%
\noindent\textbf{Enslaved.org Representations}
% \label{subsubsec:enslaved-rep}
%%%%%
The original Enslaved KG \cite{shimizu2020enslaved,enslaved} also uses a modular approach to representing agents and their various characteristics, which include race, age, sex, and occupation in addition to other factors like events and locations. Because it is more developed comparatively and therefore possesses additional characteristics, we reduced the schema to a portion that was present in the alternative version to reduce bias in results. Our motivation was that having additional concepts integrated into one representation but not the other could potentially yield unequivalent outcomes due simply to that fact, rather than the nature of the representations and an LLM's ability to understand them. This alternative Enslaved KG representation, Enslaved Wikibase, uses the software underlying Wikidata to construct the Enslaved KG \cite{shimizu2024ontology}. Since an ecosystem of supporting software and documentation already exists for Wikibase, its use for the construction of a KG would define the extent to which an already existing framework could be used for its development and deployment.

%%%%%%%%%%%%%%%%%%%%%%%%%%%%%%%%%%%%%%%%%%%%%%%%%%%%%%%%%%%%%%%%%
\subsection{CQs and Their Complexity}
\label{subsec:cqs-qc}
%%%%%
We obtained the CQs for this experiment from existing sources for KWG-Lite~\cite{site:kwg-construct-queries} and Enslaved~\cite{site:enslaved-competency-questions}, consisting of various queries, such as SPARQL CONSTRUCT queries, which we modeled as CQs. To evaluate the effects of different CQs on SPARQL query generation, we curated our selection to specific ones, classifying them based on their complexity. Specifically, we focused on three categories: simple, which requires only single hops among classes and access to the most basic data in the KG; moderate, which requires multiple but no more than three hops and some operations like simple joins and filters on the extracted data; and complex, which requires more than three hops and operations like intricate joins and filters. To limit bias toward the representations to which the CQs correspond, we also generalized them, giving the LLM more liberty in the inference of classes and relationships based on the provided context.

%%%%%%%%%%%%%%%%%%%%%%%%%%%%%%%%%%%%%%%%%%%%%%%%%%%%%%%%%%%%%%%%%
\subsection{LLM-Based Query Generation}
\label{subsec:llm-qg}
%%%%%
In the actual generation of SPARQL queries using LLMs, we considered three models: Mistral Large \cite{site:mistral-large}, Gemini Flash 2.0 \cite{site:gemini-flash-2.0}, and OpenAI GPT-4o \cite{site:gpt-4o}. After some analysis, we prioritized the use of GPT-4o due to the queries being generated by it more accurately capturing the semantics of the provided schemas and CQs. 

\begin{figure}[!h]
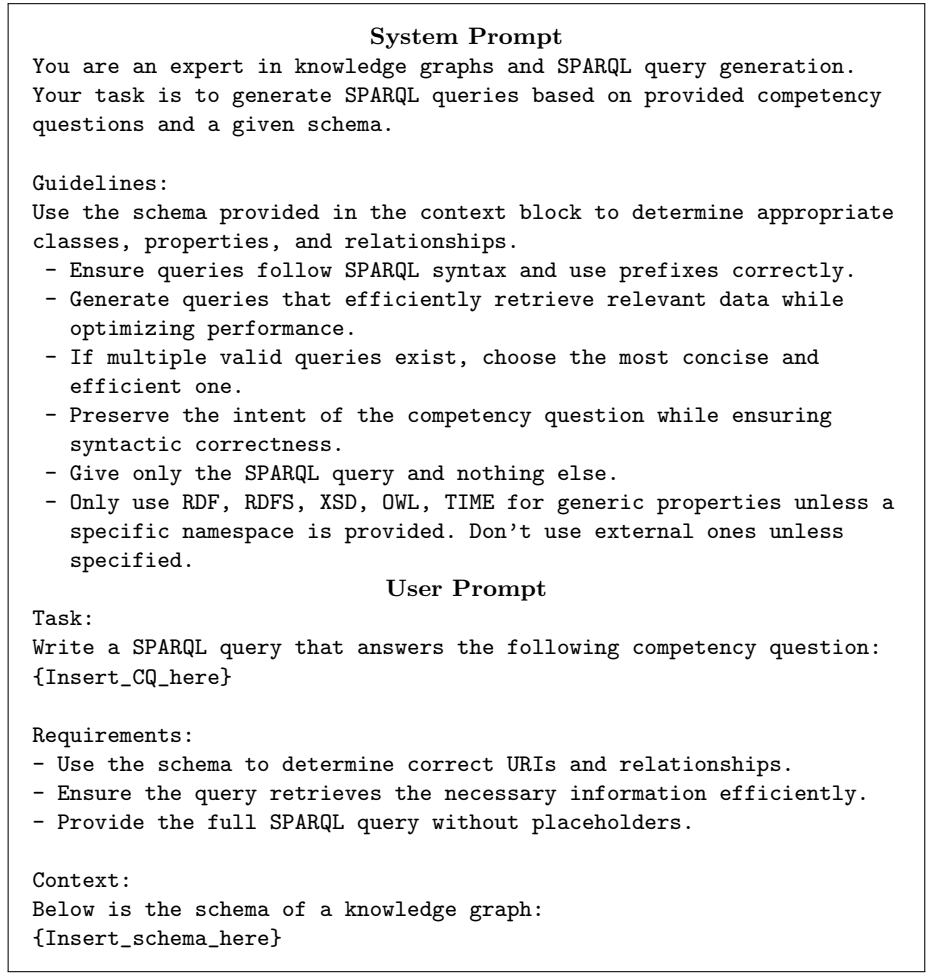

\centering
\begin{framed}
{\footnotesize
\textbf{System Prompt}
\begin{adjustwidth}{0.0cm}{0.0cm}
\begin{verbatim}
You are an expert in knowledge graphs and SPARQL query generation.
Your task is to generate SPARQL queries based on provided competency
questions and a given schema.

Guidelines:
Use the schema provided in the context block to determine appropriate
classes, properties, and relationships.
 - Ensure queries follow SPARQL syntax and use prefixes correctly.
 - Generate queries that efficiently retrieve relevant data while
   optimizing performance.
 - If multiple valid queries exist, choose the most concise and
   efficient one.
 - Preserve the intent of the competency question while ensuring
   syntactic correctness.
 - Give only the SPARQL query and nothing else.
 - Only use RDF, RDFS, XSD, OWL, TIME for generic properties unless a
   specific namespace is provided. Don't use external ones unless
   specified.
\end{verbatim}
\end{adjustwidth}
}

\textbf{User Prompt}
{\footnotesize
\begin{adjustwidth}{0.0cm}{0.0cm}
\begin{verbatim}
Task:
Write a SPARQL query that answers the following competency question:
{Insert_CQ_here}

Requirements:
- Use the schema to determine correct URIs and relationships.
- Ensure the query retrieves the necessary information efficiently.
- Provide the full SPARQL query without placeholders.

Context:
Below is the schema of a knowledge graph:
{Insert_schema_here}
\end{verbatim}
\end{adjustwidth}
}
\end{framed}
\caption{System and User Prompts}
\label{fig:prompts}
\end{figure}

In the generation process, we used a system prompt to establish basic guidelines for the LLM to follow. Additionally, we used a minimal user prompt to briefly specify the task at hand as well as provide the LLM with the requisite context (i.e., the CQ and the textual representation of the schema) in the appropriate fields in an attempt to maximize the significance of the schemas. We used a temperature of 0.8, selected after multiple iterations of sample experiments to balance creativity with consistency. The prompts used are displayed in Figure~\ref{fig:prompts}.

To make the analysis phase easier, we chose to output the CQs, final prompts, generated SPARQL queries, and raw LLM responses for each experiment to distinct spreadsheet files.

%%%%%%%%%%%%%%%%%%%%%%%%%%%%%%%%%%%%%%%%%%%%%%%%%%%%%%%%%%%%%%%%%
\subsection{Evaluation Strategy}
\label{subsec:eval}
%%%%%
For the evaluation phase, we recorded two measurements for each SPARQL query documented in the Excel files: a score reflecting its accuracy and annotations about key points of interest, specifically regarding errors, that support the chosen score and capture greater detail about its efficacy. The score is attributed based on a three-point Likert scale with \emph{-1} indicating a query with major flaws either syntactically or semantically, \emph{0} indicating minor or easily remediated flaws, and \emph{1} indicating a syntactically and semantically valid query. We decided not to penalize the queries for missing namespaces and prefixes due to the lack of their consistent presence in the provided schema information. The distribution of these scores based on CQ complexity for each representation and their respective generalized qualitative annotation summary, a mapping of which is provided in Table~\ref{table:qual-metrics}, are depicted in Table~\ref{table:results}.

%%%%%%%%%%%%%%%%%%%%%%%%%%%%%%%%%%%%%%%%%%%%%%%%%%%%%%%%%%%%%%%%%
\section{Results}
\label{sec:results}
%%%%%
%%%%%% Choice of LLMs and their stat %%%%%%%
We evaluated a set of CQs with the same schema across different LLMs -- so as to minimize confounding factors -- to identify the best model in understanding, interpreting, and generating valid SPARQL queries. OpenAI GPT-4o performed better than Gemini Flash 2.0 and Mistral Large in constructing structurally sound SPARQL queries from natural language. As such, we proceeded with GPT-4o for all subsequent experiments.

%%%%%
\begin{table}[t]
\centering
\renewcommand{\arraystretch}{1.5}
\begin{tabular}{>{\raggedright\arraybackslash}p{2.8cm}|p{8cm}@{}}
Metric & Description \\
\midrule
Clear, Consistent & Query is easy to read, well-structured, and clearly aligned with the CQ. Answers are complete and easy to trace\\
Generally Sound, Minor Gaps & Query makes sense overall and mostly works, but has small gaps in clarity, coverage, or structure that could be improved\\
Partially Complete, Schema-Tied & Query partially works, but depends too much on specific schema/CQ structures or produces incomplete answers\\
Rich, but Hard to Trace & Query appears complex or over-specified, making it difficult to verify if the answer is correct or reproducible. Results are unclear or mixed\\
Unclear, Structurally Misaligned & Query shows poor structure, unclear logic, or inconsistent alignment with the CQ, making it ineffective or misleading.\end{tabular}
\caption{Query Evaluation Annotations and Descriptions}
\label{table:qual-metrics}
\end{table}
%%%%%

%%%%%
\begin{table}[pt]
\renewcommand{\arraystretch}{1}
\renewcommand{\tabularxcolumn}[1]{>{\centering\arraybackslash}m{#1}}
\begin{tabularx}{\textwidth}{|P{1.5cm}|P{1.5cm}|P{1.3cm}|c|P{2cm}|X|}
\hline
\textbf{KG Schema} & \textbf{CQ Complexity} & \textbf{Method} & \textbf{\# CQs} & \textbf{Quantitative} & \textbf{Qualitative Summary} \\ \hline

\multirow{6}{*}{KWG} & \multirow{2}{*}{Simple} & Axioms & \multirow{2}{*}{6} & -0.67 ± 0.75 (5, 0, 1) & Partially Complete, Schema-Tied \\ \cline{3-3} \cline{5-6}
 &  & \textbf{NEN} &  & \textbf{0.50 ± 0.50 (0, 3, 3)} & \textbf{Partially Complete, Schema-Tied} \\ \cline{2-6}
 & \multirow{2}{*}{Moderate} & Axioms & \multirow{2}{*}{11} & -0.73 ± 0.62 (9, 1, 1) & Partially Complete, Schema-Tied \\ \cline{3-3} \cline{5-6}
 &  & \textbf{NEN} &  & \textbf{0.55 ± 0.50 (0, 5, 6)} & \textbf{Generally Sound, Minor Gaps} \\ \cline{2-6}
 & \multirow{2}{*}{Complex} & Axioms & \multirow{2}{*}{7} & -1.00 ± 0.00 (7, 0, 0) & Unclear, Structurally Misaligned \\ \cline{3-3} \cline{5-6}
 &  & \textbf{NEN} &  & \textbf{0.00 ± 0.00 (0, 7, 0)} & \textbf{Partially Complete, Schema-Tied} \\ \hline

\multirow{6}{*}{KWG-Lite} & \multirow{2}{*}{Simple} & \textbf{Axioms} & \multirow{2}{*}{6} & \textbf{0.83 ± 0.37 (0, 1, 5)} & \textbf{Clear, Consistent} \\ \cline{3-3} \cline{5-6}
 &  & NEN &  & 0.67 ± 0.47 (0, 2, 4) & Generally Sound, Minor Gaps \\ \cline{2-6}
 & \multirow{2}{*}{Moderate} & Axioms & \multirow{2}{*}{11} & 0.45 ± 0.50 (0, 6, 5) & Generally Sound, Minor Gaps \\ \cline{3-3} \cline{5-6}
 &  & \textbf{NEN} &  & \textbf{0.82 ± 0.39 (0, 2, 9)} & \textbf{Generally Sound, Minor Gaps} \\ \cline{2-6}
 & \multirow{2}{*}{Complex} & Axioms & \multirow{2}{*}{7} & 0.29 ± 0.70 (1, 3, 3) & Generally Sound, Minor Gaps \\ \cline{3-3} \cline{5-6}
 &  & \textbf{NEN} &  & \textbf{1.00 ± 0.00 (0, 0, 7)} & \textbf{Clear, Consistent} \\ \hline

\multirow{6}{*}{Enslaved} & \multirow{2}{*}{Simple} & \textbf{Axioms} & \multirow{2}{*}{6} & \textbf{-0.17 ± 0.69 (2, 3, 1)} & \textbf{Partially Complete, Schema-Tied} \\ \cline{3-3} \cline{5-6}
 &  & NEN &  & -0.50 ± 0.50 (3, 3, 0) & Unclear, Structurally Misaligned \\ \cline{2-6}
 & \multirow{2}{*}{Moderate} & \textbf{Axioms} & \multirow{2}{*}{9} & \textbf{-0.23 ± 0.63 (3, 5, 1)} & \textbf{Partially Complete, Schema-Tied} \\ \cline{3-3} \cline{5-6}
 &  & NEN &  & -0.33 ± 0.47 (3, 6, 0) & Partially Complete, Schema-Tied \\ \cline{2-6}
 & \multirow{2}{*}{Complex} & \textbf{Axioms} & \multirow{2}{*}{9} & \textbf{-0.33 ± 0.47 (3, 6, 0)} & \textbf{Partially Complete, Schema-Tied} \\ \cline{3-3} \cline{5-6}
 &  & NEN &  & -0.67 ± 0.47 (6, 3, 0) & Unclear, Structurally Misaligned \\ \hline

\multirow{6}{*}{\makecell{Enslaved\\Wikibase}} & \multirow{2}{*}{Simple} & \textbf{Axioms} & \multirow{2}{*}{6} & \textbf{0.34 ± 0.75 (1, 2, 3)} & \textbf{Generally Sound, Minor Gaps} \\ \cline{3-3} \cline{5-6}
 &  & NEN &  & -1.00 ± 0.00 (6, 0, 0) & Unclear, Structurally Misaligned \\ \cline{2-6}
 & \multirow{2}{*}{Moderate} & \textbf{Axioms} & \multirow{2}{*}{9} & \textbf{0.00 ± 0.82 (3, 3, 3)} & \textbf{Partially Complete, Schema-Tied} \\ \cline{3-3} \cline{5-6}
 &  & NEN &  & -0.44 ± 0.68 (5, 3, 1) & Partially Complete, Schema-Tied \\ \cline{2-6}
 & \multirow{2}{*}{Complex} & \textbf{Axioms} & \multirow{2}{*}{9} & \textbf{-0.23 ± 0.63 (3, 5, 1)} & \textbf{Partially Complete, Schema-Tied} \\ \cline{3-3} \cline{5-6}
 &  & NEN &  & -0.67 ± 0.67 (7, 1, 1) & Unclear, Structurally Misaligned \\ \hline

\end{tabularx}
\caption{Table of Results}
\smallskip
\parbox{\textwidth}{
\small \textit{Note:} 
\begin{enumerate}
    \item Values in the Quantitative column represent $mean(\mu) \pm standard deviation(\sigma)$ and the frequency distribution across evaluation classes: not accurate (-1), partially accurate (0), and accurate (1), respectively.
    \item Values bolded in a row indicate the best performer within that comparison.
\end{enumerate}
}
\label{table:results}
\end{table}

%%%%%%%%%%%%%%%%%%%%%%%%%%%%%%%%%%%%%%%%%%%%%%%%%%%%%%%%%%%%%%%%%
\subsection{KWG Results}
\label{subsec:kwg-res}
%%%%%
The queries generated for the NEN representation had performance that varied with schema complexity. On the original KWG schema, the model generally conveyed the purpose of the CQs but often prioritized the extraction of some data over the most relevant data, especially as the complexity increased. A majority of the queries predominantly appear to be well-formed with occasional minor issues such as an incorrect class or relationship. As such, of the 24 queries, 9 were correct, 15 partially correct, and none incorrect~($\mu = 0.38,~\sigma = 0.48$). In contrast, the KWG-Lite NEN performed substantially better~($\mu = 0.83,~\sigma = 0.37$), with 20 correct and 4 partially correct queries, showing improved and accurately modeled classes and relationships and more visible inferences, supporting \emph{H1}. For the axiomized representations, the model again performed reasonably better on KWG-Lite (13 correct, 10 partial, and 1 incorrect; $\mu = 0.50,~\sigma = 0.58$) than on KWG (2 correct, 1 partial, and 21 incorrect; $\mu = -0.79,~\sigma = 0.58$). Issues like usage of nonexistent classes or properties or misinterpreted relationships in the ontology arose, showing that KWG's increased complexity posed significant challenges and that the simpler, cleaner schema helped the model (supporting \emph{H1}).

In terms of CQ complexity, the discrepancy in NEN performance noticeably increased as the CQs became more complex and started a trend showing comparable results in the beginning stage. However, they diverged, with KWG diminishing toward the end and KWG-Lite expanding in performance across simple, moderate, and complex (KWG: all partial; KWG-Lite: all correct) CQs, further supporting the \emph{H1} hypothesis. Axiomatic representations showed a clearer separation. KWG-Lite consistently helped the model do reasonably well in generating structurally and semantically sound queries across all CQ complexities, while KWG struggled, with queries often missing key classes or relationships entirely and logic failing to hold. This highlights that as the cognitive load increases, imposed by both question and structure complexity, KWG-Lite's simpler structure makes a big difference in keeping the model on track.

Overall, these findings interestingly show that axiomatization may help with simpler or smaller schemas, but schema simplicity and clarity are more critical in generating solid queries (supporting \emph{H1} and partially \emph{H2}).

%%%%%%%%%%%%%%%%%%%%%%%%%%%%%%%%%%%%%%%%%%%%%%%%%%%%%%%%%%%%%%%%%
\subsection{Enslaved Results}
\label{subsec:enslaved-res}
%%%%%

Comparing the model's ability to generate queries with NEN representations of the Enslaved and Enslaved Wikibase schemas, the model appeared to perform better on the Enslaved schema~($\mu = -0.50,~\sigma = 0.50$) than on Enslaved Wikibase~($\mu = -0.67,~\sigma = 0.62$) but produced more fully correct results on Wikibase despite a higher rate of hallucinations. For Enslaved, 13 of the queries were incorrect, 11 partially correct, and 0 correct, while for Enslaved Wikibase, 18 were incorrect, 4 partially correct, and 2 correct. Further high-level analysis of the queries generated using the axiomatization method of representation showed similar overall performance, with Wikibase~($\mu = 0.00,~\sigma = 0.76$) performing slightly better than the original~($\mu = -0.25,~\sigma = 0.60$). When analyzed in terms of completeness and accuracy, Wikibase achieved more fully accurate results~(n~=~7, ~29.17\%) compared to the original~(n~=~2, ~8.34\%). Additionally, upon analysis of the method, axiomatization did better for smaller representations (Wikibase) compared to larger (Enslaved), particularly in terms of reduced hallucinations and better link resolution, helping the model hop between entities using correct predicates (rejecting \emph{H2}).

By CQ complexity, most queries generated over the Enslaved NEN schema had a solid underlying query structure, but the model's hallucinations rendered the queries only partially correct. Concerning the Enslaved Wikibase schema, the queries for simple questions were all incorrect. When the model could not find relevant information in the schema, it often substituted unrelated schema elements or hallucinated new fictitious relationships that do not exist in the schema by mixing and matching any existing ones. This is also observed with moderate and complex questions, where the model struggled in a similar vein to ground the queries correctly across both schema structures. In contrast, axiomatization on Enslaved Wikibase performed better across all CQ levels. For simple CQs, the only consistent issues were with managing prefixes, while the syntactic structure was always held strongly and tightly with the CQ, which was not the case for Enslaved where missed or incorrect relationship chaining and hallucinations were predominant. The trend continued at the moderate level, with a minute reduction in tightness with the CQ, which for Enslaved, further increased its missing predicate reasoning and hallucinations. For complex CQs, with the trend persisting with more hallucinations, performance decreased significantly for both, especially in attempting to retrieve nonexistent data from the schema to force an answer according to the CQ.

Overall, these findings suggest that while Wikibase does experience hallucinations, it still supports more complete and accurate query generation than the larger Enslaved schema, even more so when axiomatized (rejecting \emph{H2}). It also highlights that schema size significantly impacts LLM performance, with smaller schemas achieving better performance.

Detailed compilation of all results can be found in Table~\ref{table:results}.

%%%%%%%%%%%%%%%%%%%%%%%%%%%%%%%%%%%%%%%%%%%%%%%%%%%%%%%%%%%%%%%%%
\section{Discussion}
\label{sec:disc}
%%%%%
Our results indicate that knowledge conceptualization and representation significantly affect RAG efficacy, although the relationship is neither linear nor uniform across all KG schema configurations. Below, we discuss several interesting patterns and trade-offs across both schema structure and expression modalities.

While results for the KWG family support \emph{H1}, which may lead to the observation that simpler schema structure leads to better query generation, the same could not be identified with the Enslaved family. Mixed results are observed, which can be argued either as contradictory results or as uncertain generation. This arises when a more pattern-heavy variant, Wikibase, is compared with its larger modular counterpart, considering both NEN and axiomatic forms of expression. This implies that schema complexity alone does not hinder an LLM's performance, but interaction between schema complexity and the expression mode affects the model's interpretability. At the same time, we can observe that not just simpler, but smaller and more expressive (more coverage across a KG) representation led to better performance.

The trade-offs between schema complexity and representation are another interesting aspect. LLMs have a finite context window\footnote{Although, this seems to be growing as the LLM ecosystem matures.}~\cite{vaswani2017attention}, suggesting that exposing larger or heavily axiomatized schemas may exceed real-world practical prompt limits. This raises an important trade-off: ``More [context] is better'' is true, but only up to a point where a model can process it effectively. Thus, it is essential to find not just an optimal method, but a balanced middle ground and ways to tune it -- such as schema summarization~\cite{liu2018graph}, partial injection, or on-demand schema selection~\cite{song2025injectingdomainspecificknowledgelarge} -- such that it is feasible to achieve in real-world cases where the knowledge structures might be very heavy. Additionally, it emphasizes the need for future RAG systems to support dynamic schema reasoning, where only relevant partial extracts are injected based on the competency question and retrieval task at hand.

Furthermore, we find discrepancies in the extent and quality of LLM inference between simple and complex schemas. In particular, the inference is more conspicuous in the use of simpler schemas, likely due to a smaller number of classes, which necessitates greater liberty in mapping various CQ concepts to existing classes. In the case of more complex representations, however, we speculate that this increased flexibility hinders accuracy due to the existence of additional classes requiring more discrete mappings and the associated finely tuned semantics. Thus, when such inferences are made for these cases, they are more likely to induce hallucinations. We posit that this trend can be attributed, at least in part, to the LLM's temperature parameter~\cite{bellemare2024divergent}. Since we chose a higher temperature value in our experiments, we contend that the heightened creativity resulted in more suitable queries for simple but not complex schemas.

Another limitation observed in the experimentation process with our representation method for axiomatization is the use of Manchester Syntax. While we initially chose it for its greater human readability, which would be seemingly ideal for LLM interpretation, it fails to completely capture the semantic richness of the ontology, especially with complex OWL constructs such as general class axioms~\cite{gehrke2023extending}. This trade-off became clearly evident when handling Enslaved Wikibase schema, where we experienced a great difficulty in expressing a few axioms, with some not able to be expressed at all. We posit that tools such as the W3C Ontology Lexicon (OntoLex) lexicon model for ontologies (lemon) could be used for forming syntactic structures more formally \cite{site:ontolex}. This limitation further amplified the challenges, revealing a broader methodological concern with the choice of the representation language and that naive assumptions about interpretability come at the cost of semantic fidelity. 

The discovery that the axiomatized Wikibase representation yielded the best results in the Enslaved family, despite the limitations noted previously, informs a broader dialogue on the role of such frameworks in KG construction and implementation. Our findings suggest that the additional structure imposed by these frameworks, especially when designed to capture the semantics of diverse datasets, assists LLMs in retrieval tasks. Thus, the continued exploration of platforms like Wikibase as an extension to existing KGs and even in the prototyping of novel schemas to be used in conjunction with LLMs holds promise for the improved acquisition of relevant knowledge.

Over multiple runs in the experiment, we observed a consistent hallucination pattern: the introduction of non-existent classes and relationships in the queries. This typically happened when the schema lacked a direct mapping to terms in the CQ, leading to fabrications of plausible-looking URIs. One potential method to mitigate this may be explicit prompt constraints~\cite{lu2023bounding} (e.g., instructing the LLM to only generate queries using entities present in the provided schema or defer when said schema mappings appear insufficient). This, along with schema validation feedback loops~\cite{boylan2024kgvalidator} or interactive refinement prompting \cite{madaan2023self}, could potentially reduce hallucinations and further improve alignment with target knowledge structures.

One of the main aspects of our positioned agentic RAG system is its autonomous nature, where it is able to select knowledge sources and retrieve from them the bits and parts of knowledge they might possess. However, the success of this system is contingent on the knowledge conceptualization choices we make. This, as discussed, may include schema size, expressivity, representation format, and prompt engineering strategies. Finding the right balance between these factors is necessary for building scalable agentic RAG systems that are beyond controlled settings, which paves a roadmap for future research and implications for system design.

%%%%%%%%%%%%%%%%%%%%%%%%%%%%%%%%%%%%%%%%%%%%%%%%%%%%%%%%%%%%%%%%%
\section{Conclusion}
\label{sec:conc}
%%%%%
In this paper, we have outlined an extensive experiment to evaluate exactly \emph{how} ontology structure (i.e., complexity) and representation (i.e., serialization) impact the ability of an LLM to formulate a correct SPARQL query. This work is a foundational component of an overall attempt to understand agentic RAG processes for enabling interpretable prompt injection during complex multi-turn conversations relying on external knowledge bases. Our results have shown that query generation for a schema is strongly influenced by both complexity and representation, but in ways that generally relate to the CQ. In particular, we notice that hallucination occurs most frequently when the CQ does not contain direct mappings to the terms in the schema, and that syntactic and semantic inaccuracies are more likely to appear when the ``hops'' between concepts do not necessarily match the language formulation.

Finally, besides those mentioned in Section~\ref{sec:disc}, there are many next steps for further innovating in this space and exploring different capabilities of generative AI systems. We posit that conducting ablation studies, along with discussed control strategies, could offer better concrete understanding of how specific schema characteristics affect LLM performance. More exploratory directions, e.g., exploring vision language model interpretation of schema diagrams instead of textual representation and examining the combination of two or more knowledge structures for selection, extraction, and retrieval tasks, can further help in understanding how knowledge is conceptualized and interpreted by AI systems.

%%%%%%%%%%%%%%%%%%%%%%%%%%%%%%%%%%%%%%%%%%%%%%%%%%%%%%%%%%%%%%%%%
% \paragraph*{Supplemental Materials Statement:}
% %%%%%
% All data is available via the supplemental files in Easy Chair, as at the time of writing and submission, Anonymous Science is down. After review, the preceding statement will be replaced with the public repository. This represents a documented repository, which includes prompts, responses, evaluation (and criteria), and pertinent documentation for requirements. 

{\fontsize{9}{11}
% NOTE-TO-SELF: DONT EVER LEAVE ACKNOWLEDGEMENTS IN DOUBLE BLIND
\noindent\textbf{Acknowledgment.} Chris Davis Jaldi, Elham Ghiasi, and Cogan Shimizu acknowledge support from the National Science Foundation under award \#2333532 ``Proto-OKN Theme 3: An Education Gateway for the Proto-OKN (EduGate).'' O. Divine Eziolise acknowledges support from the NSF LSAMP program. Anmol Saini acknowledges support from DAGSI RX24-22. Any opinions, findings, and conclusions or recommendations expressed in this material are those of the author(s) and do not necessarily reflect the views of National Science Foundation.
Icons in Figures~\ref{fig:positioning} and \ref{fig:exps} are designed by Freepik.\smallskip

\noindent\textbf{Disclosure of Interests.} The authors have no competing interests to disclose.
}

%%%%%%%%%%%%%%%%%%%%%%%%%%%%%%%%%%%%%%%%%%%%%%%%%%%%%%%%%%%%%%%%%
%%%%%%%%%%%%%%%%%%%%%%%%%%%%%%%%%%%%%%%%%%%%%%%%%%%%%%%%%%%%%%%%%
% \bibliographystyle{splncs04}
% \bibliography{refs}

%%%%%%%%%%%%%%%%%%%%%%%%%%%%%%%%%%%%%%%%%%%%%%%%%%%%%%%%%%%%%%%%%
%%%%%%%%%%%%%%%%%%%%%%%%%%%%%%%%%%%%%%%%%%%%%%%%%%%%%%%%%%%%%%%%%
\end{document}